\documentclass[11pt]{article}

\usepackage[final]{acl}

\newcommand{\method}{\textsc{Detect--Remask--Repair}}
\newcommand{\maskdisc}{\textsc{[Mask]-Disc}}

\usepackage[normalem]{ulem}
\usepackage[table]{xcolor}
\definecolor{repairgreen}{RGB}{0,120,0}

\newcommand{\bad}[1]{\textcolor{red}{\textbf{#1}}}
\newcommand{\fix}[1]{\textcolor{repairgreen}{\textbf{#1}}}

\newif\ifshowhao
\showhaofalse
\long\def\hao#1{\ifshowhao{\color{red}#1}\else #1\fi}

\usepackage[normalem]{ulem} 
\newcommand{\haodel}[1]{\ifshowhao{\textcolor{red}{\sout{#1}}}\fi}

\usepackage{times}
\usepackage{latexsym}
\usepackage[T1]{fontenc}
\usepackage[utf8]{inputenc}
\usepackage{microtype}
\usepackage{inconsolata}
\usepackage{graphicx}
\usepackage{algorithm}
\usepackage{algpseudocode}
\usepackage{amsmath}
\usepackage{amsfonts}
\usepackage{booktabs}
\usepackage{multirow}
\usepackage{colortbl}
\usepackage{xcolor}
\usepackage{graphicx}
\definecolor{lightgray}{gray}{0.92}
\usepackage{multirow}

\title{Detect, Remask, Repair: Diffusion Editing \\for Faithful Summarization of Evolving Contexts}

\author{
Hao Zou, Zachary Horvitz, Chandhru Karthick, Zhou Yu, Kathleen McKeown \\
Columbia University \\
New York, NY, USA \\
\texttt{\{hz2999,zfh2000,ck3255,zy2461\}@columbia.edu} \\
\texttt{kathy@cs.columbia.edu}
}

\begin{document}

\maketitle


\begin{abstract}
Summaries of real-world events can become outdated as contexts evolve and new information arrives. 
A common solution is to generate an entirely new summary from the updated context. However, full regeneration discards the previous draft, can obscure what changed, and may be unnecessary when only a few claims are unsupported. We study \textbf{localized faithfulness repair}: updating outdated spans in an existing summary while preserving supported content. We propose \textbf{\method{}}, a diffusion-based framework that identifies, remasks, and repairs outdated regions with masked diffusion language models. To evaluate evolving-context summarization, we introduce \textbf{StreamSum}, a benchmark of synthetic event timelines. 
Experiments on DialogSum and StreamSum show that localized diffusion repair provides a controllable alternative to fully regenerating summaries. Faithfulness-steered repair improves stale drafts while preserving more of the original summary, and one-step variants add less than half a second of repair latency. Our results across methods illustrate trade-offs between faithfulness, latency, and preservation.
We also show that the framework can be applied as a post-hoc correction step to autoregressive outputs.
\end{abstract}

\section{Introduction}

\begin{figure}[t]
    \centering
    \includegraphics[width=\columnwidth]{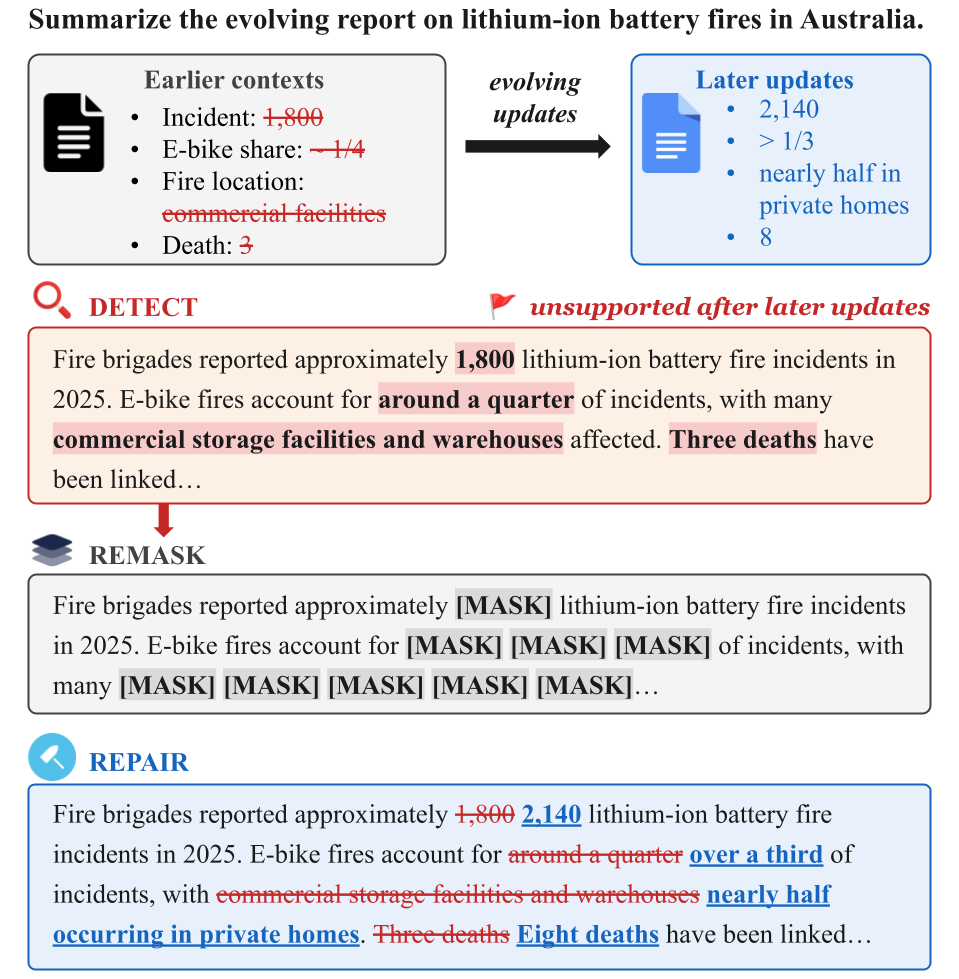}
    \caption{
    Overview of \method{} (\textsc{DRR}) on a StreamSum example. 
    In an evolving context, an autoregressive draft generated from earlier updates preserves outdated claims. 
    \textsc{DRR} uses a masked diffusion language model to first detect potentially outdated spans, then to selectively re-mask these regions, and then to infill the masked positions based on the later updates, yielding a repaired summary.
    }
    \label{fig:overview}
\end{figure}

Existing work on abstractive summarization generally assumes a static setting, where summaries are generated from fixed source documents. In contrast, real-world summarization settings are often dynamic. In breaking news, financial reports, public-safety incidents, and meetings, information arrives over time. Summaries generated from partial contexts may initially be plausible but become stale as later evidence materializes. Figure~\ref{fig:overview} illustrates this setting: claims in an early summary about lithium-ion battery fires become unsupported after later updates revise the incident count, affected locations, and death toll. 

A natural solution to evolving contexts is to regenerate an entire summary given each update. While effective, full regeneration may alter already-correct content, obscure changes, and add cost when only a subset of facts change. This is undesirable when compute is limited, or when summaries are persistent artifacts that may be displayed to users, show updates over time,  edited by humans, or consumed by downstream systems. 
We therefore study \textbf{localized faithfulness repair}: given an existing summary and an updated context, identifying and revising only unsupported spans while otherwise preserving supported content.

\begin{figure}[t]
    \centering
    \includegraphics[width=\columnwidth]{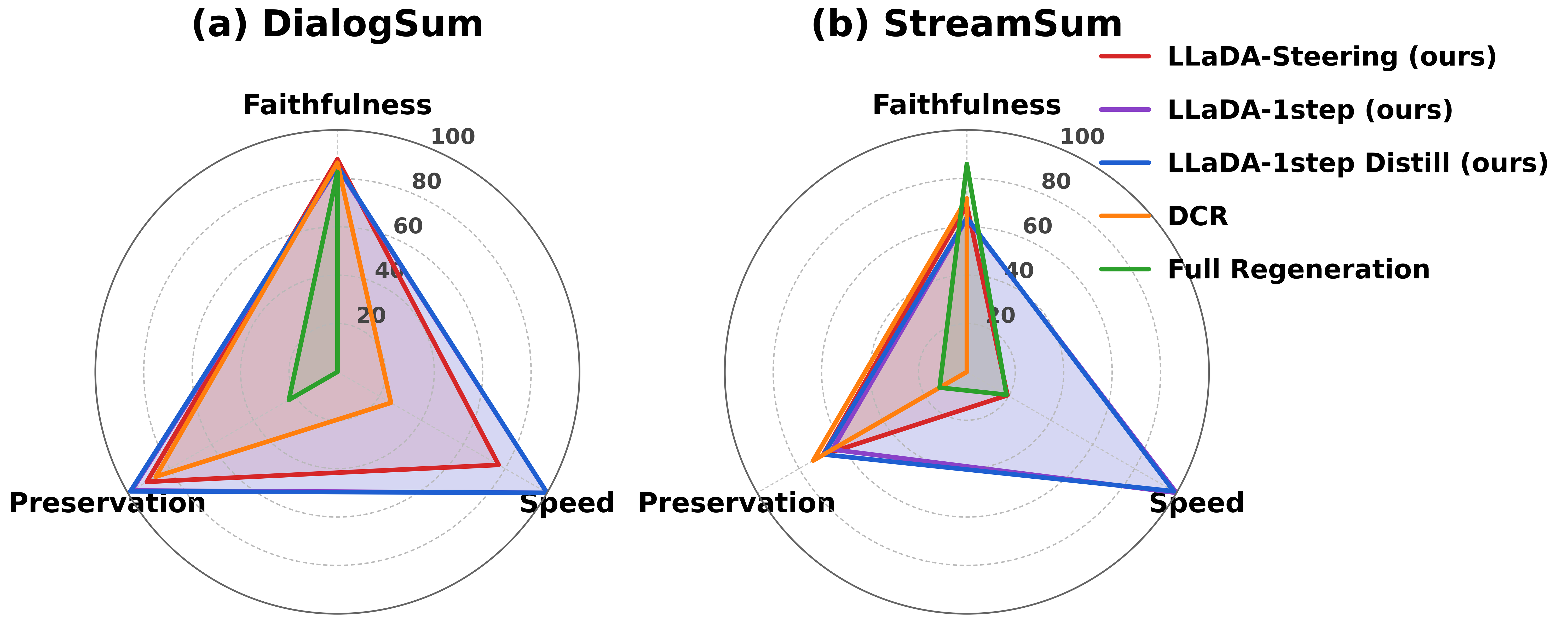}
    \caption{
Metrics for early-context summary repair methods. 
Repair methods differ in faithfulness, speed, and preservation of the existing summary. Faithfulness is measured by AlignScore~\citep{zha-etal-2023-alignscore}, preservation by inverse normalized edit distance, and speed by a normalized runtime score where higher values indicate faster methods.
Values are normalized within each dataset for the visualization.
On DialogSum, LLaDA-Steering provides the highest faithfulness while preserving much of the draft. On StreamSum, one-step methods are fastest, while full regeneration gives the highest faithfulness but lower preservation.
}
    \label{fig:operating_profile}
\end{figure}

Figure~\ref{fig:operating_profile}  shows the tradeoffs studied in this paper. 
Because the input is an existing summary, repair methods are evaluated not only by final faithfulness, but also by how much content is rewritten and latency. 
We therefore compare the different versions of summarization repair developed in this paper along three axes: faithfulness, preservation of the existing summary, and speed. 
Depending on the task, a system may prioritize rapid updates, higher faithfulness, or minimal changes to already-supported content.

To perform localized faithfulness repair, we propose \textbf{\method{}}, a diffusion-based framework for identifying stale spans and repairing them in-place.
\haodel{Our method first applies a token-level detector trained on samples from the diffusion model's unmasking process to identify incorrect summary tokens; we refer to as \textbf{\maskdisc{}}.}
\hao{Our method first applies a token-level detector, which we call \textbf{\maskdisc{}}, trained on samples from the diffusion model's unmasking process to identify summary tokens that may require repair.}
We then selectively re-mask those tokens and repair them by generating new tokens using a masked diffusion language model conditioned on the updated context. This design uses the infilling capability of masked diffusion models: instead of rewriting left-to-right, the model regenerates only selected spans while conditioning on both the source and surrounding summary context.

To evaluate localized faithfulness repair, we also introduce \textbf{StreamSum}, a benchmark for evolving-context summarization. StreamSum contains synthetic event timelines seeded by real-world news, where early summaries become stale after later reports come in, which  may overturn preliminary claims, shift responsibility or attribution across actors, or change the status of an event. 
Unlike standard static summarization datasets, StreamSum directly evaluates whether systems can accurately repair outdated claims
under evolving evidence.

We conduct experiments on StreamSum, as well as another dataset, DialogSum \cite{chen-etal-2021-dialogsum} which contains summaries of multi-turn conversations. 
Experiments on both datasets show that diffusion repair improves faithfulness of {\em early-context summaries} (i.e., summaries generated before all dialogue turns or event updates are available). 
Additionally, diffusion repair provides controllable faithfulness--speed--preservation tradeoffs. 
On DialogSum, our approach significantly improves AlignScore over both initial early-context summaries and an autoregressive critique-and-revise baseline \cite{Wadhwa2024LearningTR}. 
On StreamSum, diffusion repair substantially improves out-of-date early-context summaries.
Beyond early-context summary repair, \method{} can also be applied as a post-hoc correction layer to full-generation outputs.
Faithfulness-steered repair significantly improves AlignScore over full regeneration on DialogSum and provides a small gain on StreamSum, while diffusion repair variants consistently improve ROUGE-L \cite{lin-2004-rouge} across post-hoc repair settings.

Our main contributions are: (1) We formulate the task of \textbf{localized faithfulness repair} for evolving-context summarization, where summaries generated from partial context must be updated as new evidence arrives. (2) We propose \textbf{\method{}}, a token-level detect--remask--repair framework that combines outdated span detection with masked diffusion repair. (3) As part of our method we propose \textbf{\maskdisc{}}, a lightweight token-level detector used for both span selection and budgeted repair routing.
(4) We introduce \textbf{StreamSum}, an evolving-event summarization benchmark constructed from real-world news seeds and synthetic data.
(5) We demonstrate that diffusion repair improves early-context summaries with controllable faithfulness--speed--preservation tradeoffs and can further improve the faithfulness of full-context summaries from autoregressive systems.

\section{Related Work}
\paragraph{Faithfulness and refinement in summarization.}
Factual inconsistency has been widely studied in abstractive summarization, where generated summaries may include unsupported entities, numbers, relations, or events \cite{tang2024tofuevalevaluatinghallucinationsllms, wan2023faithfulnessawaredecodingstrategiesabstractive, maynez-etal-2020-faithfulness, pagnoni2021understandingfactualityabstractivesummarization, kryscinski-etal-2020-evaluating, fabbri2021summevalreevaluatingsummarizationevaluation, fabbri2022qafactevalimprovedqabasedfactual, wan2025positionalbiasfaithfulnesslongform}. 
Prior work addresses this problem through factuality metrics and detectors 
\cite{kryscinski-etal-2020-evaluating, goyal2020evaluating, scialom2021questevalsummarizationasksfactbased, zha-etal-2023-alignscore, tang-etal-2024-minicheck}, as well as training objectives, reranking, post-editing, decoding constraints, and critique-based refinement \cite{nan2021entitylevelfactualconsistencyabstractive, wan2023faithfulnessawaredecodingstrategiesabstractive, madaan2023selfrefineiterativerefinementselffeedback, Wadhwa2024LearningTR}. 
Our work is closely related to Detect--Critique--Refine (DCR) \cite{Wadhwa2024LearningTR}, which decomposes factual correction into detecting problematic sentences, generating natural-language feedback, and refining the output with an autoregressive model. Rather than producing sentence-level critiques and rewriting autoregressively, we use a masked diffusion model to identify and directly edit incorrect tokens.

\paragraph{Diffusion language models and remasking.}
Masked diffusion language models generate text by iteratively unmasking tokens, offering an alternative to left-to-right autoregressive generation with any-order, infilling, and parallel token decoding capabilities \cite{Li2022DiffusionLMIC, Sahoo2024SimpleAE, nie2025large, horvitz2026rethinkingreasoningmdlmsearly, zou2023surveydiffusionmodelsnatural}. Recent work also explores remasking to improve diffusion generation: \citet{wang2025remasking} introduces inference-time remasking for pretrained masked diffusion models,
\citet{peng2026pathplanningmaskeddiffusion} proposes Path Planning, which selects tokens to update at each step and enables refinement of previously unmasked tokens,
and \citet{huang2025don} trains diffusion models to identify and remask incorrect tokens using randomly masked or randomly replaced text. 
In contrast, we perform targeted remasking using \maskdisc{}, a classifier trained on model-refilled summary corruptions which contain fluent but potentially unsupported tokens.

\paragraph{Summarizing evolving contexts.} Prior work on update summarization, timeline summarization, and incremental summarization studies how to select or generate summaries from evolving document streams \cite{hwang-etal-2024-enhancing, aslam2013trec, hu-etal-2024-moments, mccreadie2014ius, zopf-etal-2016-sequential}. 
These settings assume the summary is generated anew from the current stream, so they neither localize the claims that later evidence invalidates nor measure how much of an existing summary is preserved.
StreamSum targets this complementary problem: each example pairs an early and a full context with a stale draft, a full-context reference, and revision-type metadata, which lets us measure edit locality, preservation of supported content, and when repair should give way to regeneration.


\section{Methods}

\begin{figure*}[ht!]
    \centering
    \includegraphics[width=1.0\textwidth]{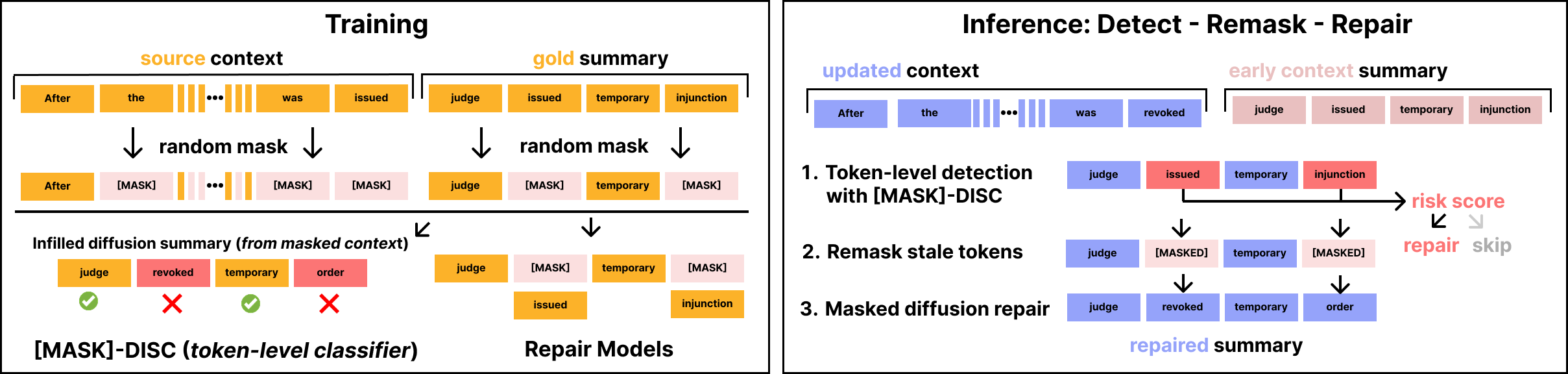}
    \caption{
    Overview of \method{}. 
During training (left), we mask the source and reference summary, then use a masked diffusion model to refill selected summary positions, producing corrupted summaries that supervise both \maskdisc{} and the one-step repair model. 
At inference (right), \maskdisc{} scores draft tokens, high-priority positions are re-masked, and a masked diffusion model infills them using the updated context while preserving supported spans. An aggregate repair-priority score can also route summaries to repair or skip.
    }
    \label{fig:method_overview}
\end{figure*}

We propose \method{}, a framework for localized summary repair under evolving contexts. 
Given a draft summary produced from partial context, the goal is to identify unsupported spans under the updated context and revise only those regions. 
This section first defines the evolving-context repair problem, then describes the \method{} pipeline. 
We then explain how \maskdisc{} is trained from diffusion-style corruptions and describe three repair variants: iterative faithfulness-steered repair, fast one-step repair, and DCR-distilled one-step repair. 
We also evaluate a  multi-sampling extension of the distilled one-step model.
Figure~\ref{fig:method_overview} summarizes the training and inference pipeline.


\subsection{Problem Setup}

We use $x_{\mathrm{early}}$ for the context available when an initial summary is produced and $x_{\mathrm{full}}$ for the updated context used for repair. 
Although our experiments use an early/full split, the same formulation can be applied to intermediate updates by treating the currently available accumulated context as $x_{\mathrm{full}}$.

In our formulation, we start with an initial summary, based on the initial context $x_{\mathrm{early}}$. For example, an  autoregressive summarizer can produce this initial summary $y_{\mathrm{early}} \sim p_{\mathrm{AR}}(y \mid x_{\mathrm{early}})$.
As new evidence arrives, some claims in $y_{\mathrm{early}}$ may become unsupported with respect to $x_{\mathrm{full}}$. 
The goal of evolving-context summary repair is to produce a repaired summary $y_{\mathrm{rep}} = R(x_{\mathrm{full}}, y_{\mathrm{early}})$
that is faithful to the updated context while preserving supported content from the draft. 
Unlike full regeneration, which rewrites a new summary from $x_{\mathrm{full}}$, localized repair aims to update only stale spans, such as revised numbers, attributions, or events, while copying supported surrounding phrasing whenever possible.

\subsection{\method{}}
\method{} decomposes localized repair into three steps. 

First, a token-level detector $D_{\theta}$ assigns each draft token an edit-priority score:
\[
s_i = D_{\theta}(x_{\mathrm{full}}, y_{\mathrm{early}}, i),
\]
where larger values indicate tokens more likely to require repair under the detector's synthetic mismatch proxy.
Second, the system selects a set of high-priority positions $M_k$ and converts them into mask tokens:
\[
y^{\mathrm{mask}}_i =
\begin{cases}
\texttt{[MASK]}, & i \in M_k,\\
y_i, & \text{otherwise}.
\end{cases}
\]
Optionally, selected tokens are expanded into short spans to enable larger edits.
Third, a repair model $G_{\phi}$ fills the masked positions conditioned on the updated context:
\[
y_{\mathrm{rep}} = G_{\phi}(x_{\mathrm{full}}, y^{\mathrm{mask}}).
\]
All unselected tokens remain fixed.
The full inference procedure is given in Appendix~\ref{app:algorithm}.

\paragraph{Selective repair routing.}
The same detector also supports sample-level repair routing.
For each draft summary, we compute a sample-level repair-priority score by averaging the top-\(k\) token scores:
\[
\rho(y) = \frac{1}{k}\sum_{i \in \operatorname{TopK}(s,k)} s_i .
\]
A selective routing policy repairs only the top \(p\%\) highest-priority summaries and skips the rest.
This lets \method{} avoid over-editing already faithful summaries.

\subsection{Training \maskdisc{} with Diffusion Unmasking}

Masked diffusion models are trained to iteratively predict \textit{masked tokens}, but they do not directly score visible tokens for whether they should be revised.
We therefore train \maskdisc{} as a lightweight token-level discriminator.
This objective is related to replaced-token detection in ELECTRA~\citep{clark2020electrapretrainingtextencoders} and to the learned per-token quality scores in \citet{kim2026finetuningmaskeddiffusionprovable}. However, \maskdisc{} is conditioned on complete source-documents and specifically trained to detect tokens generated using corrupted source-documents.

\maskdisc{} is implemented as a lightweight token-level classifier on top of masked diffusion language model representations, using a linear classification head to score visible summary tokens for repair.
Because human token-level hallucination labels are expensive, we construct synthetic supervision from diffusion-style corruptions. 
Starting from a faithful context-summary pair $(x,y_0)$, we sample a noise level and apply it to both the source context and reference summary, producing a masked context $\tilde{x}$ and a partially masked summary. 
We then run a masked diffusion language model on $\tilde{x}$ concatenated with the partially masked summary, but only refill selected masked summary positions. 
This yields a corrupted summary $y_t$ containing gold tokens, model-filled tokens, and remaining masks. 
Because the refiller conditions on a partially masked context, the filled tokens can be fluent and plausible while still unsupported by the original source.
Notably, token mismatch with the reference is only a proxy for factual unsupportedness, since faithful paraphrases may differ from the reference.
However, random-mask controls provide evidence that this proxy is useful for edit localization in practice (Section~\ref{sec:results}; Appendix~\ref{app:mask_selection_controls}).

We label each visible token by comparing it with the reference:
\[
z_i = \mathbb{I}[y_{t,i}=y_{0,i}],
\qquad
m_i = \mathbb{I}[y_{t,i}\neq \texttt{[MASK]}].
\]
Remaining mask tokens are ignored. 
The detector predicts
\[
p_i = p_{\theta}(z_i=1 \mid x,y_t,i)
\]
and minimizes token-level cross entropy:
\[
\mathcal{L}_{\mathrm{disc}}
=
-\sum_{i=1}^{L} m_i
\left[
z_i \log p_i + (1-z_i)\log(1-p_i)
\right].
\]
We define the edit-priority score as \(s_i = 1-p_i\), so higher values indicate tokens more likely to require repair under this proxy.
At inference time, we use these scores for re-masking and example-level routing.

\subsection{Repair Models}

The repair module receives the updated context $x_{\mathrm{full}}$ and selectively masked summary $y^{\mathrm{mask}}$, then fills the masked spans while leaving unmasked tokens fixed. 
\maskdisc{} determines \emph{which} spans are re-masked and, when budgeted routing is enabled, \emph{which} examples are repaired. The repair model determines \emph{how} the selected masks are filled.
We instantiate this module with three variants.




\paragraph{Iterative faithfulness-steered repair.}
Our main iterative variant uses LLaDA \cite{nie2025large} to fill the spans selected by \maskdisc{} through masked diffusion decoding. Only the selected spans are allowed to change. To improve faithfulness, we add FK-Steering~\cite{Singhal2025AGF} with BS-Fact, implemented as BERTScore \cite{zhang2020bertscoreevaluatingtextgeneration} precision against $x_{\mathrm{full}}$, as the source-grounded reward. During decoding, particles are scored by comparing estimated clean summaries against the full context and resampled toward higher-reward trajectories. Thus, \maskdisc{} supplies edit locations and routing, while FK-Steering guides the denoising process after masks are selected. Further details are in Appendix~\ref{app:fk_details}.


\paragraph{Fast one-step repair.}
The one-step variants use the same \maskdisc{} detection, re-masking, and optional routing procedure, but replace iterative masked diffusion decoding with a single repair pass.
Given the updated context and selectively masked summary, the one-step model predicts repaired tokens:
\[
q_{\psi}(y_0\mid x_{\mathrm{full}},y^{\mathrm{mask}}).
\]
The model is trained on the same corrupted-to-clean pairs used for repair supervision, with loss applied to selected repair positions. This compresses localized repair to one model evaluation and yields a low-latency repair variant. The full objective is given in Appendix~\ref{app:onestep_details}.


\paragraph{DCR-distilled repair.}
We also build on Detect--Critique--Refine (DCR)~\cite{Wadhwa2024LearningTR}, a strong autoregressive refinement framework that detects factual errors, generates natural-language feedback, and revises the summary. Instead of using DCR directly at inference time, we use its refinements as teacher targets for one-step masked repair. This distills a strong autoregressive refiner into a faster masked repair model while preserving the same explicit re-masking step used by \method{}.

\paragraph{Time-matched multiple sampling.}
For DCR-distilled repair, we also evaluate a time-matched multi-sampling extension.
The time budget is set from the autoregressive DCR baseline: for each dataset, we first measure DCR's average per-example repair time, then use that value as the per-example wall-clock cap for sampling one-step diffusion repair candidates.
This gives the distilled one-step model approximately the same average repair-time budget as DCR, but lets it spend that budget on multiple fast candidate repairs.
We then select the candidate with the highest AlignScore against the updated source context.
This is a decoding-time extension of the distilled one-step model.
\section{Experiments}
\label{sec:experiments}

We evaluate whether \method{} can repair summaries when later evidence changes the support for earlier claims. 
Our experiments are designed to answer four questions: 
(1) Can localized diffusion repair improve early-context summaries, 
(2) How does the proposed method compare with autoregressive critique-and-revise refinement, 
(3) What quality--efficiency tradeoffs arise from iterative and one-step repair, and 
(4) Can diffusion repair also serve as a post-hoc corrector for full-context summaries? 
Dataset statistics, timing, and hardware details are provided in Appendix~\ref{app:dataset_stats} and Appendix~\ref{app:timing_details}.

\subsection{Datasets and Evaluation Settings}

\paragraph{DialogSum.}
We evaluate on DialogSum~\cite{chen-etal-2021-dialogsum}, a dialogue summarization benchmark. 
Dialogues provide a natural setting for evolving-context repair: in meetings, interviews, or customer-support conversations, a system may produce an interim summary before the conversation has ended, and later turns can add details, clarify earlier statements, or change the interpretation of prior claims.
To simulate this setting, we split each dialogue into an early context and a full context.
A LLaMA-3-8B \cite{llama3} autoregressive summarizer first generates a draft summary from the early context. 
Repair methods then receive the draft and the full context, and are asked to correct unsupported or incomplete claims while preserving supported content. 
This provides a controlled setting for testing localized repair when only part of the source was initially available.

\paragraph{StreamSum.}
We also evaluate on \textbf{StreamSum}, our evolving-event summarization benchmark. 
Each example contains an early context, a full updated context, and a full-context reference summary. 
StreamSum covers six broad event domains --- politics, business, disasters, international affairs, sports, and science/technology --- and targets update patterns where later evidence changes the support for earlier claims, including numeric revisions, attribution changes, status changes, timeline changes, location changes, and outcome reversals.
To construct StreamSum, we use an agentic synthesis pipeline seeded by real-world articles retrieved from NewsDataHub\footnote{\url{https://www.newsdatahub.com/}} through API calls. 
Synthesis agents generate revision schemas, instantiate multi-update event timelines, and write full-context reference summaries, while verifier and pruning agents filter examples for realism, revision clarity, and diversity.
To ensure that later updates are necessary, we use AlignScore to compare how well the full-context reference summary is supported by the early context versus the full context, retaining examples with weak early support and strong full-context support.
We use the train and validation splits to train \maskdisc{} and one-step repair models, and evaluate on the held-out test split. 
Additional construction details, prompts, dataset statistics, and representative examples are provided in Appendix~\ref{app:streamsum_generation}.

\paragraph{Repair settings.}
We evaluate two settings. In \textbf{early-context summary repair},  
$x_{\mathrm{early}}$ is the first half of the source context and $x_{\mathrm{full}}$ is the complete source context, and systems repair summaries generated from $x_{\mathrm{early}}$ using $x_{\mathrm{full}}$ as evidence. In \textbf{post-hoc full-generation repair}, systems receive a summary already generated from $x_{\mathrm{full}}$ and attempt to correct remaining unsupported spans.
The first setting tests summary generation under evolving evidence, while the second tests whether the same \method{} framework can serve as a post-hoc faithfulness correction step.

\subsection{Baselines and Systems}

We compare against three main baselines. \textbf{AR Draft} is generated by LLaMA-3-8B from the early context and measures the quality of summaries before later evidence is available. \textbf{Full Regeneration} prompts LLaMA-3-8B to fully generate an entirely new summary directly from the full context.  
\textbf{DCR} is an autoregressive detect--critique--refine baseline~\cite{Wadhwa2024LearningTR}, which detects errors, generates natural-language feedback, and refines the summary using a fine-tuned autoregressive model.

We evaluate three repair models within \method{}: \textbf{LLaDA-Steering}, \textbf{LLaDA-1step}, and \textbf{LLaDA-1step Distill}. These methods differ in the repair model used to fill the selected masks. All three use \maskdisc{} for span selection and optional example-level routing. LLaDA-Steering uses iterative masked diffusion repair with FK-Steering and a BS-Fact source-support reward computed against $x_{\mathrm{full}}$. The one-step variants use a single-pass masked repair model, with the distilled version trained from DCR teacher refinements.

\subsection{Evaluation Metrics}
We report ROUGE-L \cite{lin-2004-rouge} and BLEURT \cite{sellam-etal-2020-bleurt} for summary quality, and AlignScore \cite{zha-etal-2023-alignscore} (AS) as our primary automatic faithfulness metric. 
Unless otherwise stated, AlignScore is always computed against the full context, including for AR drafts generated from early context.
\hao{
ROUGE-L is computed against the official DialogSum reference summaries for DialogSum and against the full-context reference summary for StreamSum.
These reference-based metrics measure similarity to reference summaries, while AlignScore measures source support against the updated context.
}
To measure preservation, we report normalized token edit distance from the input draft. Lower values indicate fewer changes to the existing summary.
Efficiency is measured by the number of function evaluations (NFE) and average wall-clock time per example. 
We assess statistical significance for AlignScore using paired bootstrap resampling over examples with 10{,}000 resamples. 
Unless otherwise noted, repair time excludes the cost of initial summary generation.
Metrics details are provided in Appendix~\ref{app:evaluation_details}.

\begin{table*}[!t]
\centering
\small
\setlength{\tabcolsep}{2.7pt}
\begin{tabular}{l|ccccc|ccccc}
\toprule
\multirow{2}{*}{\textbf{Method}}
& \multicolumn{5}{c|}{\textbf{DialogSum}}
& \multicolumn{5}{c}{\textbf{StreamSum}} \\
\cmidrule(lr){2-6} \cmidrule(lr){7-11}
& \textbf{R-L}$\uparrow$ & \textbf{BLEURT}$\uparrow$ & \textbf{AS}$\uparrow$ & \textbf{NFE}$\downarrow$ & \textbf{Time}$\downarrow$
& \textbf{R-L}$\uparrow$ & \textbf{BLEURT}$\uparrow$ & \textbf{AS}$\uparrow$ & \textbf{NFE}$\downarrow$ & \textbf{Time}$\downarrow$ \\
\midrule

\multicolumn{11}{@{}l}{\textit{AR Draft (early context)}} \\
LLaMA-3-8B
& 0.1628 & 0.5058 & 0.8513 & 59.92 & 1.61
& 0.2261 & 0.4908 & 0.5432 & 117.07 & 3.20 \\

\addlinespace[3pt]
\multicolumn{11}{@{}l}{\textit{Diffusion Repair (ours)}} \\
\rowcolor{gray!12}
\textbf{LLaDA$_{\text{steering}}$}
& \textbf{0.1866} & 0.5055 & \textbf{0.8790}$^{\dagger\ddagger}$ & 7.84 & 0.89
& \textbf{0.2515} & 0.4795 & 0.6895$^{\ddagger}$ & 32.00 & 7.69 \\

\rowcolor{gray!12}
\textbf{LLaDA$_{\text{1-step}}$}
& 0.1752 & 0.4867 & 0.8474 & \textbf{0.25} & \textbf{0.04}
& 0.2500 & 0.3403 & 0.6345$^{\ddagger}$ & \textbf{1.00} & \textbf{0.31} \\

\rowcolor{gray!12}
\textbf{LLaDA$_{\text{1-step Distill}}$}
& 0.1746 & 0.4908 & 0.8518 & \textbf{0.25} & \textbf{0.04}
& 0.2462 & 0.3505 & 0.6377$^{\ddagger}$ & \textbf{1.00} & 0.43 \\

\rowcolor{gray!12}
\quad + MS (AlignScore rerank)
& 0.1750 & 0.4927 & \textbf{0.8664}$^{\ddagger}$ & 15.19 & 2.68
& 0.2480 & 0.3490 & \textbf{0.7843}$^{\dagger\ddagger}$ & 27.64 & 9.28 \\

\addlinespace[3pt]
\multicolumn{11}{@{}l}{\textit{Autoregressive Refinement}} \\
DCR (LLaMA-3-8B)
& 0.1618 & 0.4980 & 0.8649 & 78.51 & 2.78
& 0.2354 & 0.4870 & 0.7174 & 277.00 & 9.46 \\

\addlinespace[3pt]
\multicolumn{11}{@{}l}{\textit{Full Regeneration (full context)}} \\
LLaMA-3-8B
& 0.1762 & \textbf{0.5318} & 0.8258 & 87.14 & 3.72
& \textbf{0.3489} & \textbf{0.5303} & \textbf{0.8596} & 127.05 & 7.73 \\

\bottomrule
\end{tabular}
\caption{
Main results for early-context summary repair. 
\hao{AS denotes AlignScore.}
Rows shaded in gray are our diffusion repair variants. 
MS denotes budgeted multi-sampling over LLaDA$_{\text{1-step Distill}}$ candidates with AlignScore reranking. 
\(\dagger\) indicates significantly better AS than DCR, and \(\ddagger\) indicates significantly better AS than the AR draft, under paired bootstrap resampling (\(p<0.05\)). 
Time for repair methods reports additional repair cost averaged over all examples, including examples that are routed to skip repair. 
}
\label{tab:main_results_dialogsum_streamsum}
\end{table*}

\subsection{Implementation Details}
\label{sec:implementation_details}


We provide full hyperparameters in Appendix~\ref{sec:appendix_experimental_details}.
Unless otherwise specified, iterative repair uses 32 denoising steps and 4 particles. 
For DialogSum, we use conservative budgeted repair by default, repairing only the highest-risk 25\% of examples, because many early-context dialogue summaries are already compatible with the full dialogue and unnecessary edits can reduce faithfulness and preservation.
For StreamSum, we use the full repair budget because later event updates more often invalidate early-context summaries and require correcting multiple related facts.
The full budget sweep is reported in Appendix~\ref{app:budgeted_repair}.
One-step repair models are trained with LoRA adapters on diffusion-style corrupted summaries.
\section{Results and Analysis}
\label{sec:results}

We evaluate \method{} along three axes: faithfulness, speed, and preservation.
Early-context summary repair refers to repairing a summary generated from the first half of the input using the full context as evidence.

\subsection{Local Repair of Early-Context Summaries}

Table~\ref{tab:main_results_dialogsum_streamsum} shows two different repair regimes. 
On DialogSum, the AR draft generated from the first half of the dialogue is already relatively faithful to the full dialogue, with an AlignScore of 0.8513. 
LLaDA-Steering with conservative repair routing improves AlignScore to 0.8790, a statistically significant improvement over both the AR draft and DCR under paired bootstrap resampling. 
This result suggests that the detector can allocate repair computation to examples that benefit from local correction while avoiding unnecessary edits to already faithful summaries.

StreamSum exhibits a different pattern. 
The AR draft generated from the first half of the evolving event context has much lower AlignScore against the full context, 0.5432, showing that later updates substantially change the support for early summaries. 
LLaDA-Steering significantly improves AlignScore to 0.6895 under paired bootstrap resampling. 
The one-step variants provide much lower-latency repair, reaching 0.6345 and 0.6377 AlignScore with only 0.31--0.43 seconds of additional repair time.

Multi-sampling substantially improves the distilled one-step model, raising AlignScore from 0.8518 to 0.8664 on DialogSum and from 0.6377 to 0.7843 on StreamSum.
On StreamSum, this also improves over DCR while using far fewer function evaluations.

Full regeneration remains strongest overall on StreamSum, reaching 0.8596 AlignScore.
This suggests that localized repair does not always suffice when later evidence substantially changes the final summary content.
Appendices~\ref{app:streamsum_revision_types}, \ref{app:llm_factuality_audit}, and \ref{app:human_factuality_audit} provide complementary evidence for this interpretation.
The human audit shows that \method{} reduces unsupported claims relative to early-context summaries and, on StreamSum, substantially reduces outdated claims.

\begin{table}[t]
\centering
\small
\setlength{\tabcolsep}{6pt}
\begin{tabular}{lcc}
\toprule
\textbf{Method} & \textbf{DialogSum}$\downarrow$ & \textbf{StreamSum}$\downarrow$ \\
\midrule
\rowcolor{gray!12}
LLaDA$_{\text{steering}}$ & 0.0909 & 0.3140 \\
\rowcolor{gray!12}
LLaDA$_{\text{1-step}}$ & 0.0192 & 0.3582 \\
\rowcolor{gray!12}
LLaDA$_{\text{1-step Distill}}$ & 0.0123 & 0.3169 \\
DCR & 0.1336 & 0.2671 \\
Full Regeneration & 0.7686 & 0.8701 \\
\bottomrule
\end{tabular}
\caption{
Preservation under early-context summary repair, measured as normalized token edit distance from the AR draft.
}
\label{tab:preservation_edit_distance}
\end{table}

\paragraph{Preservation.}
Preservation matters in evolving-context summarization because the input summary may already be displayed to users, reviewed by humans, or consumed by downstream systems; unnecessary rewriting can obscure what changed and introduce new errors.
Table~\ref{tab:preservation_edit_distance} reports normalized token edit distance from the early-context summary. 
Full regeneration rewrites substantially more of the draft than repair methods on both datasets. 
LLaDA-Steering achieves the best DialogSum AlignScore with relatively small edits. On StreamSum, diffusion repair provides faster targeted corrections that preserve more text, while full regeneration attains higher faithfulness through larger edits.
These results illustrate the faithfulness--speed--preservation tradeoff in Figure~\ref{fig:operating_profile}.

\paragraph{Localized-editing controls.}
\begin{table}[t]
\centering
\small
\setlength{\tabcolsep}{5pt}
\begin{tabular}{lcccc}
\toprule
\textbf{Setting} & \textbf{AS}$\uparrow$ & \textbf{\(\Delta\)AS}$\uparrow$ & \textbf{Edit}$\downarrow$ & \textbf{Time}$\downarrow$ \\
\midrule
AR Draft & 0.8513 & -- & 0.0000 & -- \\
AR same-mask & 0.8619 & +0.0106 & 0.1582 & 2.45s \\
\rowcolor{gray!12}
DRR (ours) & \textbf{0.8790}$^{\dagger}$ & \textbf{+0.0277} & \textbf{0.0909} & \textbf{0.89s} \\
\bottomrule
\end{tabular}
\caption{\hao{
Same-mask AR infilling control on DialogSum.
LLaMA-3.1-8B-Instruct fills the identical \maskdisc{} spans used by DRR.
AS denotes AlignScore and Edit is normalized token edit distance from the AR draft.
\(\dagger\) indicates significantly better AS than AR same-mask under paired bootstrap resampling (\(p<0.05\)).
}}
\label{tab:same_mask_ar}
\end{table}

To isolate the effect of masked diffusion repair from the selected edit locations, Table~\ref{tab:same_mask_ar} compares DRR with a separate same-mask AR infilling baseline. It gives LLaMA-3.1-8B-Instruct the identical \maskdisc{}-selected spans used by DRR and asks it to infill those spans, holding the edit locations fixed.
The LLaMA same-mask baseline improves over the draft, which is consistent with \maskdisc{} providing useful edit localization.
Under identical edit locations, DRR achieves a larger faithfulness gain with less editing and lower latency.
Additional random-mask controls in Appendix~\ref{app:mask_selection_controls} further show that, although \maskdisc{}-selected spans are imperfect proxies for factual errors, they provide a useful signal for deciding where repair should be applied.

\begin{table*}[t]
\centering
\small
\setlength{\tabcolsep}{2.7pt}
\begin{tabular}{l|ccccc|ccccc}
\toprule
\multirow{2}{*}{\textbf{Method}}
& \multicolumn{5}{c|}{\textbf{DialogSum}}
& \multicolumn{5}{c}{\textbf{StreamSum}} \\
\cmidrule(lr){2-6} \cmidrule(lr){7-11}
& \textbf{R-L}$\uparrow$ & \textbf{BLEURT}$\uparrow$ & \textbf{AS}$\uparrow$ & \textbf{NFE}$\downarrow$ & \textbf{Time}$\downarrow$
& \textbf{R-L}$\uparrow$ & \textbf{BLEURT}$\uparrow$ & \textbf{AS}$\uparrow$ & \textbf{NFE}$\downarrow$ & \textbf{Time}$\downarrow$ \\
\midrule

\multicolumn{11}{@{}l}{\textit{Full Regeneration (full context)}} \\
LLaMA-3-8B
& 0.1762 & \textbf{0.5318} & 0.8258 & 87.14 & 3.72
& 0.3489 & 0.5303 & 0.8596 & 127.05 & 7.73 \\

\addlinespace[3pt]
\multicolumn{11}{@{}l}{\textit{Diffusion Repair (ours)}} \\
\rowcolor{gray!12}
\textbf{LLaDA$_{\text{steering}}$}
& \textbf{0.1844} & 0.5306 & 0.8355$^{\ddagger}$ & 31.97 & 10.46
& \textbf{0.3625} & \textbf{0.5423} & \textbf{0.8638} & 32.00 & 7.61 \\

\rowcolor{gray!12}
\textbf{LLaDA$_{\text{1-step}}$}
& 0.1789 & 0.5282 & 0.8241 & \textbf{1.00} & \textbf{0.18}
& 0.3602 & 0.4704 & 0.8507 & \textbf{0.25} & \textbf{0.04} \\

\rowcolor{gray!12}
\textbf{LLaDA$_{\text{1-step Distill}}$}
& 0.1775 & 0.5290 & 0.8304 & \textbf{1.00} & 0.19
& 0.3608 & 0.3369 & 0.8530 & \textbf{0.25} & 0.08 \\

\addlinespace[3pt]
\multicolumn{11}{@{}l}{\textit{Autoregressive Refinement}} \\
DCR (LLaMA-3-8B)
& 0.1667 & 0.5310 & \textbf{0.8378} & 140.48 & 4.96
& 0.3594 & 0.5340 & 0.8632 & 16.83 & 0.78 \\

\bottomrule
\end{tabular}
\caption{
Post-hoc repair of full-generation outputs. 
\(\ddagger\) indicates significantly better AlignScore than the full-generation input under paired bootstrap resampling (\(p<0.05\)). 
NFE and time for diffusion repair report additional repair cost averaged over all examples.
}
\label{tab:posthoc_fullgen}
\end{table*}

\subsection{Effect of Budgeted Repair}
Budgeted repair is motivated by the fact that not every early-context summary needs correction.
If a draft is already compatible with the full context, rewriting it can unnecessarily change supported content or introduce new errors.
We therefore use \maskdisc{} as a sample-level router: examples are ranked by an aggregate token-level repair score, and only the top \(p\%\) highest-risk examples are repaired.
This creates a controllable setting where the system can focus repair on summaries most likely to benefit, while preserving already-supported summaries and reducing computation.
Appendix~\ref{app:budgeted_repair} reports the full Top-25/50/75/All results and timing breakdowns.

The trends differ across datasets.
On DialogSum, conservative routing is most effective: LLaDA-Steering reaches 0.8790 AlignScore when repairing only the top 25\% highest-risk examples, but drops to 0.8584 when repairing all examples.
This suggests that many early-context dialogue summaries are already compatible with the full dialogue, so unnecessary edits can hurt faithfulness.
On StreamSum, repair improves as the budget increases: LLaDA-Steering rises from 0.5856 at Top-25 to 0.6895 when all examples are repaired.
This matches the benchmark design, where later event updates more often make early summaries outdated.
Together, these results suggest that \maskdisc{} can prioritize repair for summaries that are more likely to benefit under a limited budget.

\subsection{Post-hoc Repair of Full Generations}

Table~\ref{tab:posthoc_fullgen} evaluates whether localized repair remains useful after full-context generation. 
Unlike early-context repair, the input summary has already been generated from the full context, so repair targets remaining unsupported spans rather than newly available evidence.

On DialogSum, LLaDA-Steering improves the AlignScore of the LLaMA full-generation from 0.8258 to 0.8355, a statistically significant gain under paired bootstrap resampling. 
The difference from DCR's AlignScore of 0.8378 is not statistically significant. 
Gains are modest because full-generation outputs are already strong, but the result shows that token-level detection and masked diffusion repair can still find useful local corrections after full-summary generation.

The post-hoc results also clarify the roles of our repair variants. 
Faithfulness-steered repair explicitly optimizes a source-grounded reward during editing and is the strongest post-hoc corrector among our diffusion variants.
The one-step variants are much faster and consistently improve ROUGE-L over full generation, but they are less reliable for AlignScore in this setting. 

\subsection{Efficient and Interpretable Repair}

The results show a tradeoff between faithfulness, computation, and the amount of text that must be rewritten. 
The one-step variants provide the strongest speed advantage, reducing repair to a single model evaluation. 
Iterative faithfulness-steered repair is slower, but provides stronger faithfulness control. Together, these results suggest a practical workflow: use \maskdisc{} to estimate repair risk, choose a repair budget based on the desired cost--preservation tradeoff, and apply either fast one-step repair or iterative steering depending on resource constraints.  Overall, these results position localized diffusion repair as a promising editing tool.

DCR is a strong autoregressive refinement baseline, but it relies on natural-language feedback and autoregressive rewriting.
In contrast, \method{} operates with in-place token-level operations: \maskdisc{} scores each summary token, selected spans are re-masked, and the repair model fills only those masks. 
This results in an interpretable pipeline where users can see which spans were selected, how many examples were routed for repair, and how changing the repair budget affects both faithfulness and cost.
Representative repair examples with edited spans highlighted are shown in Appendix~\ref{app:repair_examples}.

When existing summaries are mostly reusable, as in DialogSum, \method{} can improve faithfulness while preserving supported text and limiting repair cost. When later updates require broader rewriting, as in StreamSum, full regeneration achieves higher faithfulness. In such cases, localized repair remains useful when explicit edit locations, preservation, or low-latency correction are part of the task requirements.
\section{Conclusion}
\label{sec:conclusion}

We introduced \method{}, a diffusion-based framework for localized faithfulness repair.
When context changes, instead of regenerating an entire summary, \method{} estimates which text to repair, identifies tokens likely to require repair, selectively re-masks them, and repairs the masked spans with masked diffusion language models. 
We also introduced StreamSum, an evolving-context summarization benchmark where early summaries become stale as later updates arrive. Our results show that localized diffusion repair can provide an efficient, controllable, and interpretable approach that can repair factual inconsistencies and complement autoregressive methods.

\clearpage
\section*{Limitations}
\label{sec:limitations}

Our evaluation relies on automatic metrics such as AlignScore, ROUGE-L, and BLEURT, which are useful but imperfect proxies for faithfulness and summary quality. 
This limitation is especially notable for StreamSum because AlignScore is also used during benchmark construction for filtering.
We supplement these metrics with revision-type analysis, an LLM-assisted factuality audit, and a small human factuality audit. However, these diagnostics are not a substitute for full-scale human evaluation.
While seeded by real-world news, StreamSum is synthetic, and future work should evaluate localized repair on naturally occurring evolving-summary data. 
Finally, while \maskdisc{} provides token-level edit candidates, the current system uses fixed edit budgets. Methods that adaptively select the number of positions to re-mask could better balance faithfulness, preservation, and speed.

\section*{Ethical Considerations}
\label{sec:ethics}

This work studies faithfulness repair for summaries in evolving contexts. 
Improving factual consistency can reduce the risk of downstream harms from users relying on stale information or unsupported claims. However, automatic repair systems should not be treated as a substitute for human verification in high-stakes domains. 
Our method exposes selected edit locations and repairs them with a masked diffusion model, but it may still miss unsupported claims or introduce new errors. 
Any deployment of summarization repair systems should include transparency about model-generated edits, careful monitoring for factual errors, and human oversight where summaries may affect real-world decisions.
\section*{Acknowledgments}
\label{sec:acknowledgments}

We thank the anonymous reviewers and meta-reviewer for their constructive feedback.
This research is based upon work supported in part by the Office of the Director of National Intelligence (ODNI), Intelligence Advanced Research Projects Activity (IARPA), via 560000C260018. The views and conclusions contained herein are those of the authors and should not be interpreted as necessarily representing the official policies, either expressed or implied, of ODNI, IARPA, or the U.S. Government. The U.S. Government is authorized to reproduce and distribute reprints for governmental purposes notwithstanding any copyright annotation therein.

\paragraph{Large Language Model Use.}
We used LLM assistants for proofreading, copyediting, wording revisions, and literature-review assistance. Additionally, we used AI coding tools for assistance with scripts and plot generation.
The authors reviewed and verified all resulting outputs.

\bibliography{main}

\clearpage
\appendix
\section{Additional Method Details}
\label{app:method_details}

\subsection{Inference Algorithm}
\label{app:algorithm}

Algorithm~\ref{alg:method_appendix} gives the full inference procedure for \method{}. 
At each round, \maskdisc{} scores draft tokens, the highest-priority positions are selected and re-masked, and the repair model infills the selected spans. 
The edit schedule $\mathcal{K}=(k_1,\ldots,k_R)$ controls how many tokens are edited at each round.

\begin{algorithm}[h]
\caption{\method{} Inference}
\label{alg:method_appendix}
\begin{algorithmic}[1]
\Require Updated context $x$, draft summary $y^{(0)}$, detector $D_{\theta}$, repair model $G_{\phi}$, edit schedule $\mathcal{K}=(k_1,\ldots,k_R)$
\Ensure Repaired summary $\hat{y}$
\State $y \gets y^{(0)}$
\For{$r=1,\ldots,R$}
    \State $s \gets D_{\theta}(x,y)$ \hfill $\triangleright$ Detect suspicious tokens
    \State $M_r \gets \textsc{SelectEdits}(s,k_r)$
    \If{$M_r=\emptyset$}
        \State \textbf{break}
    \EndIf
    \State $y^{\mathrm{mask}} \gets \textsc{ReMask}(y,M_r)$ \hfill $\triangleright$ Re-mask selected spans
    \State $\tilde{y} \gets G_{\phi}(x,y^{\mathrm{mask}})$ \hfill $\triangleright$ Repair masked spans
    \State $y \gets \textsc{Merge}(y^{\mathrm{mask}},\tilde{y},M_r)$
\EndFor
\State \Return $y$
\end{algorithmic}
\end{algorithm}

\subsection{Diffusion-Style Corruption Details}
\label{app:corruption_details}

To train \maskdisc{} and the repair models, we construct corrupted summaries that resemble intermediate states of masked diffusion decoding. 
Given a reference summary $y_0$, we sample a corruption level and mask a subset of summary positions $A_t$. 
A masked diffusion language model refills a subset $B_t \subseteq A_t$, leaving the remaining positions masked. 
The resulting sequence $y_t$ contains three token types: original gold tokens, model-filled tokens, and unfilled \texttt{[MASK]} tokens.

Visible tokens are labeled by exact match with the reference:
\[
z_i = \mathbb{I}[y_{t,i}=y_{0,i}],
\qquad
m_i = \mathbb{I}[y_{t,i}\neq \texttt{[MASK]}].
\]
Tokens that remain masked are excluded from the detector loss. 
In practice, we sample multiple corruption levels to expose the detector to easy and difficult states, ranging from lightly corrupted summaries to highly masked summaries. 
This creates fluent negative examples without requiring human token-level annotations.

\subsection{Faithfulness-Steered Diffusion Repair}
\label{app:fk_details}

Let $c=(x_{\mathrm{full}}, y^{\mathrm{mask}})$ denote the repair condition. 
A discrete masked diffusion repair process defines
\[
p_{\phi}(y_{0:T}\mid c)
=
p_{\mathrm{prior}}(y_T)
\prod_{t=T}^{1}
p_{\phi}(y_{t-1}\mid y_t, c),
\]
where $y_T$ contains the selected masks and $y_0$ is the repaired summary. 
During repair, unmasked draft tokens are fixed and only selected masked positions are updated.

To encourage faithful repairs, we adapt FK-Steering \cite{Singhal2025AGF}. 
Let $r(y,x_{\mathrm{full}})$ be a source-grounded faithfulness reward. 
We target an exponentially tilted distribution:
\[
p_{\mathrm{target}}(y_0\mid x_{\mathrm{full}})
\propto
p_{\phi}(y_0\mid c)
\exp(\lambda r(y_0,x_{\mathrm{full}})).
\]
Since direct sampling from this distribution is intractable, we approximate it with particle-based decoding. 
At selected denoising steps, each particle forms an estimated clean summary $\hat{y}_0$ by filling remaining masks, evaluates $r(\hat{y}_0,x_{\mathrm{full}})$, and resamples particles according to the resulting potentials. 
This shifts probability mass toward denoising trajectories that yield more source-supported repairs.

\subsection{One-Step and DCR-Distilled Repair Objectives}
\label{app:onestep_details}

The one-step repair model predicts clean summary tokens from a corrupted summary in a single forward pass:
\[
q_{\psi}(y_0\mid x_{\mathrm{full}},y_t).
\]
We train it with token-level cross entropy over supervised repair positions:
\[
\mathcal{L}_{\mathrm{1step}}
=
-\sum_{i=1}^{L} a_i
\log q_{\psi}(y_{0,i}\mid x_{\mathrm{full}},y_t,i),
\]
where $a_i$ indicates the positions included in the repair loss. 
In our main setting, $a_i=1$ for masked positions and $0$ otherwise, so the model learns to fill selected repair spans while preserving visible tokens.

For DCR-distilled repair, we use teacher refinements $y_{\mathrm{DCR}}$ from Detect--Critique--Refine \cite{Wadhwa2024LearningTR}. 
The one-step repair model is trained to imitate these targets:
\[
\mathcal{L}_{\mathrm{distill}}
=
-\sum_i a_i
\log q_{\psi}(y_{\mathrm{DCR},i}\mid x_{\mathrm{full}},y_t,i).
\]
This provides a fast repair model trained from a stronger autoregressive refinement teacher.

\section{Experimental Details}
\label{sec:appendix_experimental_details}

\paragraph{Data construction for \maskdisc{}.}
For \maskdisc{} training, we generate diffusion-style corrupted summaries from reference context--summary pairs. 
We sample a noise level, mask both the source context and the reference summary, and run LLaDA on the masked context concatenated with the masked summary. 
Only masked summary positions are editable during refill; masked source tokens remain part of the noisy conditioning context. 
This produces partial summaries containing gold tokens, model-refilled tokens, and remaining masks. 
Visible summary tokens are labeled as correct or incorrect by exact match against the reference summary, while remaining masks are ignored.

\paragraph{Corruption hyperparameters.}
We sample denoising steps from \(\{8,16,32\}\), use a maximum step cap of 64, and sample fill fractions from \(\{0.25,0.5,0.75\}\). 
We require each corrupted example to contain at least one visible token and one remaining mask. 
The detector is trained as a token-level classifier and selected by validation F1 on incorrect-token detection.

\paragraph{Repair hyperparameters.}
Unless otherwise specified, iterative repair uses 32 denoising steps, 4 particles, and steering weight \(\lambda=6.0\). 
For DialogSum, the default token edit budget is 8 for DISC-gated repair. 
For StreamSum, the default token edit budget is 64 because summaries are longer and evolving-event updates often require correcting multiple related facts. 
One-step repair models are trained with LoRA adapters, supervising masked summary positions unless otherwise stated.

\paragraph{Budgeted DISC-gating.}
For budgeted repair, we rank examples by the mean of their top-\(k\) \maskdisc{} repair-priority scores. 
We then repair only the top \(p\%\) highest-risk examples and keep the remaining drafts unchanged. 
DialogSum uses \(k=8\), while StreamSum uses \(k=64\), matching the token-level edit budgets used by the corresponding repair pipelines.

\paragraph{Timing.}
Unless otherwise stated, repair or refinement time reports incremental correction latency after the existing draft is available.
End-to-end latency for a pipeline that first generates a draft and then repairs it is the sum of draft-generation time and repair/refinement time.
For DISC-gated settings, time is averaged over all examples, including skipped examples. 
For post-hoc full-generation repair, total pipeline time is the full-generation time plus the additional correction time.

\section{Dataset Statistics}
\label{app:dataset_stats}

\begin{table}[t]
\centering
\small
\setlength{\tabcolsep}{5pt}
\begin{tabular}{llrrr}
\toprule
\textbf{Subset} & \textbf{Dataset} & \textbf{Size} & \textbf{Doc Len} & \textbf{Sum Len} \\
\midrule
Train & DialogSum & 12460 & 194 & 32 \\
Val & DialogSum & 500 & 192 & 30 \\
Test & DialogSum & 1500 & 200 & 26 \\
\midrule
Train & StreamSum & 2000 & 458 & 92 \\
Val & StreamSum & 500 & 456 & 93 \\
Test & StreamSum & 500 & 465 & 93 \\
\bottomrule
\end{tabular}
\caption{
Dataset statistics. 
Lengths are average token counts using simple tokenization. 
For DialogSum, document length refers to the dialogue length; for StreamSum, document length refers to the full evolving context length.
}
\label{tab:dataset_statistics}
\end{table}

Table~\ref{tab:dataset_statistics} reports dataset sizes and average lengths. 

\section{Timing and Hardware Details}
\label{app:timing_details}

All timing experiments were run on NVIDIA RTX A6000 GPUs with 48GB memory. 
The machine used NVIDIA driver version 595.71.05 and CUDA 13.2. 
Timing reports wall-clock seconds per example averaged over the evaluated split. 
For repair methods, unless otherwise stated, reported time is additional repair/refinement cost and includes examples skipped by budgeted routing. 
For autoregressive LLaMA-based generation and refinement baselines, we use KV caching during decoding. 
Masked diffusion repair does not use autoregressive KV caching.

\section{Evaluation Details}
\label{app:evaluation_details}

\paragraph{Metrics.}
We evaluate summary quality with ROUGE-L and BLEURT, and evaluate faithfulness with AlignScore. 
ROUGE-L measures lexical overlap with the reference summary, while BLEURT measures learned semantic similarity to the reference. 
AlignScore measures whether a generated summary is supported by the corresponding source context, and is used as our primary automatic faithfulness metric. 
We follow the official setup and instructions for all metrics. 
For ROUGE-L, we use the ROUGE package from Google Research.\footnote{\url{https://github.com/google-research/google-research/tree/master/rouge}} 
For BS and BS-Fact, we use the default English model, RoBERTa-Large. 
All reported model scores are from single runs.

\paragraph{Preservation metric.}
We measure preservation using normalized token edit distance from the AR draft. 
Given tokenized draft \(y_{\mathrm{ar}}\) and output \(\hat{y}\), we compute
\[
\mathrm{EditDist}(y_{\mathrm{ar}}, \hat{y})
=
\frac{\mathrm{Lev}(y_{\mathrm{ar}}, \hat{y})}
{\max(|y_{\mathrm{ar}}|, |\hat{y}|)}.
\]
Lower values indicate stronger preservation of the original draft.

\paragraph{Statistical significance.}
We assess statistical significance for AlignScore using paired bootstrap resampling over test examples with 10{,}000 resamples. 
Since systems are evaluated on the same examples, we compare paired per-example AlignScore differences. 
For each bootstrap resample, we sample examples with replacement and compute the mean difference between two systems. 
A difference is considered significant when the 95\% bootstrap confidence interval excludes zero, corresponding to \(p<0.05\).

\paragraph{Efficiency measurement.}
We report the number of function evaluations (NFE) and average wall-clock time per example.
For repair and refinement methods, reported time is incremental correction latency after the existing input summary is available.
For DISC-gated repair, time is averaged over all examples, counting skipped examples as zero additional repair cost.
For full regeneration, time reports generation from the full updated context.
End-to-end latency for a pipeline that first generates an early-context summary and then repairs it is the sum of early-context generation time and incremental repair time.
For post-hoc full-generation repair, total pipeline time is the full-generation time plus the additional correction time.

\section{StreamSum Revision-Type Analysis}
\label{app:streamsum_revision_types}
We further break down StreamSum results by major revision type.
The listed categories cover 486 of the 500 test examples; the remaining 14 examples contain less frequent revision patterns.
For each type, we report AlignScore for the AR early-context summary, DRR, DCR, and full regeneration, along with the DRR improvement over the AR draft and normalized edit distance from the AR draft.

\begin{table*}[t]
\centering
\small
\setlength{\tabcolsep}{4.2pt}
\begin{tabular}{lrrrrrrrr}
\toprule
\textbf{Type} & \textbf{n} & \textbf{AR} & \textbf{DRR} & \textbf{DCR} & \textbf{Full} & \textbf{DRR \(\Delta\)} & \textbf{DRR Edit} & \textbf{Full Edit} \\
\midrule
Numeric & 206 & .5309 & .6914 & .7070 & .8528 & +.1606 & .3539 & .8930 \\
Outcome flip & 235 & .5563 & .6911 & .7275 & .8633 & +.1347 & .3622 & .9089 \\
Attribution & 19 & .5175 & .6466 & .6995 & .8775 & +.1291 & .3938 & .9060 \\
Market/financial & 16 & .5089 & .6955 & .7332 & .8448 & +.1866 & .4324 & .8878 \\
Status & 10 & .5328 & .6876 & .6458 & .9379 & +.1548 & .3966 & .8844 \\
\bottomrule
\end{tabular}
\caption{\hao{
StreamSum results by major revision type.
AR, DRR, DCR, and Full report AlignScore.
DRR \(\Delta\) is the AlignScore improvement of DRR over the AR early-context summary.
Edit is normalized token edit distance from the AR draft; lower values indicate stronger preservation.
}}
\label{tab:streamsum_revision_types}
\end{table*}

\hao{
Table~\ref{tab:streamsum_revision_types} shows that DRR consistently improves the stale AR draft across major revision types.
The largest gains appear for concrete factual substitutions, such as market/financial changes, numeric updates, and status changes.
Outcome flips and attribution changes also improve, but these cases often require coordinated changes across several summary claims.
Full regeneration remains strongest on StreamSum because many examples require broader event-level rewriting rather than isolated span repair.
This supports our scoped conclusion: localized repair is useful for targeted correction and preservation-sensitive settings, while full regeneration is preferable when later evidence changes the overall event interpretation.
}

\section{LLM-Assisted Factuality Audit}
\label{app:llm_factuality_audit}

\hao{
To supplement AlignScore-based evaluation, we conduct an LLM-assisted factuality audit.
We use GPT-5.6-Terra as the auditor with a fixed rubric and deterministic decoding, and apply the same rubric to all systems within each dataset.
The auditor is given the updated context, the draft or repaired summary, and a rubric that separately scores factual support, stale or outdated content, newly introduced unsupported information, preservation of supported content, update incorporation, and whether the summary is fully supported.
This audit is intended as an additional diagnostic rather than a replacement for human evaluation.
}
Fact., Pres., and Update are 0--2 ordinal scores where higher is better; Stale and New err. are counts of stale/outdated claims and newly introduced unsupported claims, respectively; Full supp. is the fraction of outputs judged fully supported.

\begin{table*}[t]
\centering
\small
\setlength{\tabcolsep}{4.0pt}
\begin{tabular}{lrrrrrr}
\toprule
\textbf{System} & \textbf{Fact.}\(\uparrow\) & \textbf{Stale}\(\downarrow\) & \textbf{New err.}\(\downarrow\) & \textbf{Pres.}\(\uparrow\) & \textbf{Update}\(\uparrow\) & \textbf{Full supp.}\(\uparrow\) \\
\midrule
AR Draft & 1.269 & 1.261 & 0.000 & 1.496 & 0.011 & 0.557 \\
AR same-mask & 0.917 & 0.787 & 0.312 & 0.744 & 0.253 & 0.213 \\
DRR & \textbf{1.563} & 0.629 & 0.328 & \textbf{1.795} & 0.544 & \textbf{0.640} \\
Full regen & 1.496 & \textbf{0.035} & 0.488 & 1.760 & \textbf{1.960} & 0.536 \\
\bottomrule
\end{tabular}
\caption{\hao{
LLM-assisted factuality audit on DialogSum.
DRR improves factual support and preservation over the AR draft and has fewer newly introduced unsupported claims than full regeneration, while full regeneration removes stale content and incorporates later context more aggressively.
}}
\label{tab:llm_audit_dialogsum}
\end{table*}

\hao{
On StreamSum, the same audit confirms the broader trend from the automatic evaluation: full regeneration is stronger when updates require substantial rewriting, but DRR still improves the early-context summary in factual support (0.356 to 0.514), stale-content reduction (1.972 to 1.364), and update incorporation (0.008 to 0.862).
}

\section{Small Human Factuality Audit}
\label{app:human_factuality_audit}

To supplement automatic metrics, we conduct a small human factuality audit on sampled DialogSum and StreamSum examples.
We randomly sample 10 examples from each dataset using a fixed seed, yielding 20 examples total.
For each example, we evaluate three anonymized system outputs: the AR early-context summary, DRR / LLaDA-Steering, and full regeneration.
The system order is shuffled independently for each example, yielding 60 anonymized summary judgments.
One author performed the annotations. System identities were hidden during annotation and revealed only after scoring for aggregation.

\paragraph{Annotation rubric.}
For each sampled output, the annotator was shown the full updated source context and an anonymized system summary.
The annotator recorded: (1) the number of unsupported factual claims, defined as claims not supported by the updated source; (2) the number of outdated claims, defined as claims contradicted, revised, or superseded by later evidence; and (3) a binary full-support label indicating whether all factual claims in the summary were supported by the updated source.
Missing information was not counted as an error unless the summary made an unsupported claim.

\begin{table*}[t]
\centering
\small
\setlength{\tabcolsep}{4.2pt}
\begin{tabular}{llccc}
\toprule
\textbf{Dataset} & \textbf{System} & \textbf{Unsup.}$\downarrow$ & \textbf{Outdated}$\downarrow$ & \textbf{Full supp.}\%$\uparrow$ \\
\midrule
\multirow{3}{*}{DialogSum}
& AR Draft & 0.30 & 0.00 & 80.0 \\
& DRR / LLaDA-Steering & 0.10 & 0.00 & 90.0 \\
& Full Regeneration & 0.10 & 0.00 & 90.0 \\
\midrule
\multirow{3}{*}{StreamSum}
& AR Draft & 4.00 & 3.70 & 0.0 \\
& DRR / LLaDA-Steering & 2.40 & 1.70 & 30.0 \\
& Full Regeneration & 0.00 & 0.00 & 100.0 \\
\bottomrule
\end{tabular}
\caption{
Small author-annotated human factuality audit over 10 DialogSum and 10 StreamSum examples.
Unsup. is the average number of unsupported factual claims.
Outdated is the average number of claims contradicted, revised, or superseded by the updated context.
Full supp. reports the percentage of summaries for which all factual claims are supported by the updated context.
}
\label{tab:human_factuality_audit}
\end{table*}

Table~\ref{tab:human_factuality_audit} supports the main conclusion from the automatic evaluation.
On DialogSum, all systems are mostly supported, and DRR reduces the average number of unsupported claims relative to the AR draft.
On StreamSum, early-context summaries contain many outdated claims because later event updates substantially revise the evidence.
DRR reduces both unsupported and outdated claims relative to the AR draft, but full regeneration is strongest on this sample, consistent with StreamSum often requiring coordinated event-level rewriting rather than isolated span repair.
This audit is intended as a small diagnostic supplement and should not be interpreted as a replacement for a full-scale human evaluation.

\section{Mask Selection Controls}
\label{app:mask_selection_controls}

\hao{
We evaluate whether the edit locations selected by \maskdisc{} are more useful than random edit locations.
This control addresses whether the gains in Table~\ref{tab:same_mask_ar} come from meaningful token-level localization or simply from giving the repair model an opportunity to rewrite part of the summary.
For each example, we keep the same routing policy, edit budget, and repair model, but replace the \maskdisc{}-selected token locations with random summary-token locations.
}

\begin{table}[t]
\centering
\small
\setlength{\tabcolsep}{4.2pt}
\begin{tabular}{lllcc}
\toprule
\textbf{Setting} & \textbf{Mask} & \textbf{Model} & \textbf{AS}$\uparrow$ & \textbf{Edit}$\downarrow$ \\
\midrule
AR Draft & -- & -- & 0.8513 & 0.0000 \\
AR$_{\text{Random}}$ & random & LLaMA & 0.8548 & 0.1726 \\
AR$_{\text{\maskdisc{}}}$ & disc & LLaMA & 0.8619 & 0.1582 \\
DRR$_{\text{Random}}$ & random & LLaDA$_{\text{steering}}$ & 0.8732 & 0.1177 \\
DRR$_{\text{\maskdisc{}}}$ & disc & LLaDA$_{\text{steering}}$ & \textbf{0.8790} & \textbf{0.0909} \\
\bottomrule
\end{tabular}
\caption{\hao{
Random-mask controls on DialogSum.
AS denotes AlignScore and Edit is normalized token edit distance from the AR draft.
Under the same routing and edit budget, \maskdisc{}-selected locations yield higher AS and lower edit distance than random locations for both autoregressive infilling and LLaDA repair.
}}
\label{tab:random_mask_controls}
\end{table}

\hao{
Table~\ref{tab:random_mask_controls} shows that random masking provides only small gains over the AR draft.
For autoregressive infilling, replacing random locations improves AS from 0.8513 to 0.8548, while replacing \maskdisc{} locations improves AS to 0.8619 and edits less.
The same pattern holds for diffusion repair: random masking reaches 0.8732 AS with 0.1177 edit distance, whereas \maskdisc{} masking reaches 0.8790 AS with only 0.0909 edit distance.
Thus, the detector is not merely increasing the amount of rewriting; it selects edit locations that are more useful for faithfulness repair and more conservative in preservation.
}

\hao{
Held-out detector results further support \maskdisc{} as an edit-localization module: incorrect-token F1 is 0.780 on DialogSum and 0.765 on StreamSum, while span-overlap F1 is 0.785 and 0.792, respectively.
}

\section{Budgeted Repair Results}
\label{app:budgeted_repair}

Table~\ref{tab:budgeted_disc_gating} reports AlignScore as we vary the repair budget.
We rank examples by aggregate \maskdisc{} repair-priority score, repair only the top \(p\%\) highest-risk examples, and leave the remaining examples unchanged.
We report \(p \in \{25,50,75,100\}\).
This evaluates whether \maskdisc{} can serve as a sample-level routing signal in addition to selecting token-level edit targets.

As expected, repair time increases with the budget: for early-context summary LLaDA-Steering, repair time grows from 0.89s to 3.54s on DialogSum and from 1.89s to 7.69s on StreamSum as \(p\) increases from Top-25 to All.
The one-step variants remain much cheaper across budgets, staying below 0.18s on DialogSum and below 0.44s on StreamSum.
Full timing breakdowns are provided in Table~\ref{tab:budget_timing_appendix}.

\begin{table*}[t]
\centering
\small
\setlength{\tabcolsep}{4.2pt}
\begin{tabular}{llcccc|cccc}
\toprule
\multirow{2}{*}{\textbf{Setting}} 
& \multirow{2}{*}{\textbf{Method}}
& \multicolumn{4}{c|}{\textbf{DialogSum}}
& \multicolumn{4}{c}{\textbf{StreamSum}} \\
\cmidrule(lr){3-6} \cmidrule(lr){7-10}
& & \textbf{Top-25} & \textbf{Top-50} & \textbf{Top-75} & \textbf{All}
& \textbf{Top-25} & \textbf{Top-50} & \textbf{Top-75} & \textbf{All} \\
\midrule

\multirow{3}{*}{\textit{Early draft}}
& LLaDA$_{\text{steering}}$
& \textbf{0.8790} & 0.8724 & 0.8680 & 0.8584
& 0.5856 & 0.6213 & 0.6580 & \textbf{0.6895} \\

& LLaDA$_{\text{1-step}}$
& \textbf{0.8474} & 0.8453 & 0.8421 & 0.8389
& 0.5723 & 0.6014 & 0.6210 & \textbf{0.6345} \\

& LLaDA$_{\text{1-step Distill}}$
& 0.8518 & 0.8535 & \textbf{0.8545} & 0.8534
& 0.5743 & 0.6007 & 0.6211 & \textbf{0.6377} \\

\midrule

\multirow{3}{*}{\textit{Full-gen post-hoc}}
& LLaDA$_{\text{steering}}$
& 0.8299 & 0.8334 & 0.8334 & \textbf{0.8355}
& 0.8613 & 0.8624 & 0.8622 & \textbf{0.8638} \\

& LLaDA$_{\text{1-step}}$
& 0.8248 & 0.8239 & \textbf{0.8257} & 0.8241
& \textbf{0.8507} & 0.8456 & 0.8364 & 0.8303 \\

& LLaDA$_{\text{1-step Distill}}$
& 0.8274 & 0.8275 & 0.8292 & \textbf{0.8304}
& \textbf{0.8530} & 0.8518 & 0.8419 & 0.8387 \\

\bottomrule
\end{tabular}
\caption{
Budgeted repair-routing ablation. 
We repair only the top \(p\%\) highest-risk examples ranked by \maskdisc{} and report AlignScore. 
Top-\(p\) repair leaves the remaining examples unchanged.
}
\label{tab:budgeted_disc_gating}
\end{table*}

\section{Additional Analysis of \maskdisc{} Repair-Priority Scores}
\label{app:disc_risk_correlation}

Section~\ref{sec:results} uses \maskdisc{} as a sample-level repair router for early-context summary repair. 
For each summary, we compute a repair-priority score by averaging the top-\(k\) token scores:
\[
S(x,y)=\frac{1}{k}\sum_{i\in \mathrm{TopK}(s,k)} s_i .
\]
We set \(k\) to match the token-level edit budget used by the corresponding repair pipeline. 
For DialogSum, \(k=8\), reflecting its shorter dialogue summaries and conservative edit budget. 
For StreamSum, \(k=64\), reflecting longer summaries and the need to revise multiple related facts in evolving-event updates. 

To evaluate whether \maskdisc{} provides a useful sample-level routing signal, we correlate \(S(x,y)\) with per-example AlignScore computed against the full context for early-context summaries.
A negative correlation indicates that summaries assigned higher repair priority by \maskdisc{} tend to be less supported by the full context.

\begin{table}[h]
\centering
\small
\setlength{\tabcolsep}{0.8pt}
\begin{tabular}{llccc}
\toprule
\textbf{Setting} & \textbf{Dataset} & \textbf{Pearson \(r\)} & \textbf{Spearman \(\rho\)} & \textbf{\(p\)} \\
\midrule
Early draft & DialogSum & -0.263 & -0.239 & $<0.001$ \\
Early draft & StreamSum & -0.285 & -0.299 & $<0.001$ \\
\bottomrule
\end{tabular}
\caption{
Correlation between sample-level \maskdisc{} repair-priority score and per-example AlignScore for early-context summaries.
}
\label{tab:disc_risk_correlation_appendix}
\end{table}

The early-context summary correlations are negative and statistically significant on both datasets, but small in magnitude: Spearman \(\rho=-0.239\) on DialogSum and \(\rho=-0.299\) on StreamSum.
This suggests that \maskdisc{} provides a coarse repair-routing signal in the expected direction, but should not be interpreted as a standalone faithfulness metric.
Our main evidence for budgeted routing is therefore the downstream repair behavior in Table~\ref{tab:budgeted_disc_gating}.

\paragraph{Budgeted Repair Timing}

Table~\ref{tab:budget_timing_appendix} reports repair time under different budgeted routing thresholds. 
Timing reports additional repair/refinement cost averaged over all examples, including examples skipped by the router.

\begin{table*}[t]
\centering
\small
\setlength{\tabcolsep}{4.2pt}
\begin{tabular}{llcccc|cccc}
\toprule
\multirow{2}{*}{\textbf{Setting}} 
& \multirow{2}{*}{\textbf{Method}}
& \multicolumn{4}{c|}{\textbf{DialogSum}}
& \multicolumn{4}{c}{\textbf{StreamSum}} \\
\cmidrule(lr){3-6} \cmidrule(lr){7-10}
& & \textbf{Top-25} & \textbf{Top-50} & \textbf{Top-75} & \textbf{All}
& \textbf{Top-25} & \textbf{Top-50} & \textbf{Top-75} & \textbf{All} \\
\midrule
\multirow{3}{*}{\textit{Early draft}}
& LLaDA$_{\text{steering}}$ & 0.894 & 1.820 & 2.714 & 3.544 & 1.892 & 3.803 & 5.747 & 7.687 \\
& LLaDA$_{\text{1-step}}$ & 0.044 & 0.089 & 0.133 & 0.175 & 0.077 & 0.154 & 0.231 & 0.309 \\
& LLaDA$_{\text{1-step Distill}}$ & 0.043 & 0.087 & 0.130 & 0.172 & 0.110 & 0.215 & 0.324 & 0.432 \\
\midrule
\multirow{3}{*}{\textit{Full-gen post-hoc}}
& LLaDA$_{\text{steering}}$ & 2.649 & 5.340 & 7.864 & 10.462 & 1.926 & 3.827 & 5.748 & 7.606 \\
& LLaDA$_{\text{1-step}}$ & 0.046 & 0.093 & 0.138 & 0.184 & 0.044 & 0.088 & 0.135 & 0.178 \\
& LLaDA$_{\text{1-step Distill}}$ & 0.047 & 0.094 & 0.139 & 0.186 & 0.079 & 0.158 & 0.238 & 0.316 \\
\bottomrule
\end{tabular}
\caption{
Additional repair time under budgeted routing thresholds. 
Times are seconds per example averaged over all examples, including skipped examples.
}
\label{tab:budget_timing_appendix}
\end{table*}

\section{Representative Repair Examples}
\label{app:repair_examples}

To illustrate the inspectability of \method{}, Table~\ref{tab:repair_examples} shows representative repairs.
We truncate the source context for space and highlight stale spans selected for repair in red and repaired content in green.
The model operates over tokenizer-level units; for readability, we display word-level spans.
These examples are intended to show the kinds of factual updates made by the repair model, rather than to provide a full qualitative evaluation.

\begin{table*}[t]
\centering
\footnotesize
\setlength{\tabcolsep}{5pt}
\begin{tabular}{p{0.16\textwidth}p{0.78\textwidth}}
\toprule
\textbf{Field} & \textbf{Example} \\
\midrule

\multicolumn{2}{l}{\textbf{Example 1: StreamSum, GambleChoice Awards}} \\
\midrule

Context excerpt &
Voting opened in early June with five public-vote categories and an original deadline of June 18. \ldots{}
A technical outage knocked the GambleChoice platform offline for nearly two days, prompting organizers to extend the deadline to June 25. \ldots{}
The gala, originally planned for Vienna, was relocated to Amsterdam. \ldots{}
VaultPlay Casino was ultimately named Best Casino at the Amsterdam ceremony. \\

AR draft with stale spans &
The 8th annual GambleChoice Awards has entered its public voting phase, with five categories open for voting: Best Casino, Best New Casino, Best Sportsbook, Best New Slot, and Best Game Provider. The voting period will remain open until \bad{June 18}, with winners to be announced at a gala ceremony in \bad{Vienna}. Preliminary data shows \bad{RoyalSpin Casino} leading in the Best Casino category, but organizers caution that the race is still wide open. The public vote will be combined with verified community member votes to determine the winners. \\

Masked repair budget &
The actual model input masks \textbf{64 tokenizer-level positions}: \texttt{[MASK]} $\times$ 64. For readability, we show the main affected spans rather than the full raw tokenizer-level masked sequence. \\

Repaired summary &
The 8th annual GambleChoice Awards began with a public voting period, with five categories open for voting: Best Casino, Best New Casino, Best Sportsbook, Best New Slot, and Best Game Provider. A technical outage extended the deadline to \fix{June 25}, and a sixth category was added later: \fix{Best Live Casino}. Early counts showed RoyalSpin Casino leading in the Best Casino category, but organizers relocated the ceremony to \fix{Amsterdam}. A strong surge of late submissions for \fix{VaultPlay Casino} ultimately saw it named Best Casino. \\

\midrule
\multicolumn{2}{l}{\textbf{Example 2: StreamSum, Nikkei AI Rally Reversal}} \\
\midrule

Context excerpt &
The Nikkei 225 briefly surpassed 42,800 points in the morning session before reversing and closing lower. \ldots{}
Nexora Semiconductors first projected a 20\% profit increase, but later issued corrected guidance calling for a 5\% profit decline. \ldots{}
Final settlement data showed that overseas investors were net sellers on the day. \\

AR draft with stale spans &
Japanese equities \bad{surged in early trading} on Wednesday, with the Nikkei 225 briefly reaching a new all-time intraday high above 42,800 points. The rally was driven by a broad advance in artificial intelligence and semiconductor stocks, led by Nexora Semiconductors, which projected a \bad{20\% profit increase} for the current fiscal year due to strong demand from data center operators. Strong \bad{foreign investor inflows}, a \bad{weaker yen}, and optimism over AI infrastructure spending also contributed to the market's \bad{positive sentiment}. \\

Masked repair budget &
The actual model input masks \textbf{64 tokenizer-level positions}: \texttt{[MASK]} $\times$ 64. For readability, we show the main affected spans rather than the full raw tokenizer-level masked sequence. \\

Repaired summary &
Japanese equities \fix{reversed early gains} on Wednesday, with the Nikkei 225 briefly reaching a new all-time intraday high of 42,800 points. The rally was reversed by an earnings revision in the AI-driven technology sector, led by Nexora Semiconductors, which forecast a 20\% profit increase earlier, but was later revised to \fix{a 5\% profit decline}. \fix{Reduced} foreign capital inflows, a \fix{stronger yen}, and a \fix{Bank of Japan clarification} also contributed to the \fix{selloff}. \\

\bottomrule
\end{tabular}
\caption{
Representative inspectable repair examples from StreamSum using LLaDA-Steering.
Context excerpts are shortened from the raw full StreamSum context, with \ldots{} marking omitted text.
We show stale factual spans in the AR draft, the repair mask budget, and the repaired output.
Red marks stale spans, while green marks salient repaired content; the highlighting is illustrative rather than an exhaustive token-level diff.
Both examples use 64 tokenizer-level \texttt{[MASK]} positions and 32 diffusion refinement steps.
}
\label{tab:repair_examples}
\end{table*}

\section{StreamSum Construction Details}
\label{app:streamsum_generation}

StreamSum is designed to evaluate summarization repair under evolving evidence. 
Each example contains an early context, a full updated context, an early-context summary, and a full-context reference summary. 
The goal is to create examples where a summary generated from the early context is plausible at the time it is written, but later becomes incomplete or partially unsupported after subsequent updates.
The benchmark is constructed from real-world news seeds retrieved from NewsDataHub\footnote{\url{https://www.newsdatahub.com/}} through API calls, then expanded into controlled multi-update event timelines using stage-specific synthesis and validation agents. Representative examples are shown in Table~\ref{tab:streamsum_examples}.

\paragraph{Pipeline overview.}
We construct StreamSum with an agentic synthesis pipeline seeded by real-world news. 
The pipeline has eight stages: 
(1) schema brainstorming, 
(2) seed-article collection from NewsDataHub, 
(3) timeline construction, 
(4) timeline hardening, 
(5) summary construction, 
(6) AlignScore-based support filtering, 
(7) verifier-agent validation, and 
(8) diversity-aware pruning. 

\paragraph{Models and tools.}
Generation-heavy stages use Claude Opus 4.6 for schema brainstorming, timeline construction, timeline hardening, and full-context summary construction. 
Verification and pruning stages use MiniMax-M2.5 to evaluate candidate timelines for realism, entity consistency, revision clarity, and diversity. 
The pipeline also calls AlignScore as an automatic support-scoring tool: for each candidate, we compute support between the full-context reference summary and each timeline prefix, producing a support trajectory from early updates to the full context. 
Verifier agents use these trajectories, along with the timeline text, to identify examples where later evidence is necessary for a faithful final summary.

\paragraph{Seed articles.}
We retrieve real-world seed articles from NewsDataHub using domain-balanced retrieval policies. 
The retrieval stage prioritizes articles with event-evolution potential, such as preliminary numbers, unresolved outcomes, investigations, attribution uncertainty, named entities, concrete dates, and later correction or update cues. 
We deduplicate articles by URL, title, and content hash, and filter low-value sources such as advertisements, recipes, coupons, and photo-gallery pages. 
The seed articles provide realistic entities, event types, and reporting style, while the synthesis pipeline controls the revision structure needed for evaluation.

\paragraph{Revision schemas.}
A schema-generation agent proposes candidate event schemas. 
Each schema specifies an event family, revision type, scenario description, dominant revision point, why the first half is misleading, the late evidence needed, and expected updates. 
We use six event families: politics, business, disaster, international, sports, and science. 
We use revision types including outcome flips, numeric updates, attribution updates, status changes, timeline changes, and location changes. 
Schemas are required to contain one dominant factual revision rather than multiple unrelated changes.

\paragraph{Timeline construction and hardening.}
Given a schema and seed article, a timeline-construction agent produces a coherent sequence of 4--5 news-style updates. 
Early updates contain plausible but incomplete information, while later updates introduce evidence that changes the correct final interpretation. 
A hardening agent then improves realism, removes redundancy, strengthens the revision arc, and checks that all updates remain anchored to the same central event. 
The resulting full context is the concatenation of all updates, while the early context consists of the first part of the timeline.

\paragraph{Full-context reference summary construction.}
A summary-construction agent writes a concise full-context reference summary. 
The prompt requires the summary to reflect the final state of the event, include the dominant revision, and avoid claims that are only supported by early updates. 
The summary is intended to be well supported by the full context but not fully supported by the early context.

\paragraph{AlignScore prefix filtering.}
We compute AlignScore between the full-context reference summary and each timeline prefix: \(t_1\), \(t_1{+}t_2\), ..., and the full timeline. 
This produces a support trajectory for each candidate. 
We retain examples where the final summary has strong support under the full context but weaker support under early prefixes. 
Concretely, candidates are preferred when the full-context AlignScore is high, the first-prefix AlignScore is low, and the full-minus-early support gap is large. 
This filtering encourages examples where later updates are necessary for a faithful summary.

\paragraph{Verifier and pruning agents.}
A verifier agent evaluates each candidate using both the textual timeline and the AlignScore prefix trajectory. 
It scores temporal coherence, entity consistency, dominant revision quality, early-context misleadingness, final-context support, and benchmark usefulness. 
Finally, a pruning agent selects a diverse subset, balancing event families, revision types, difficulty levels, and near-duplicate patterns.

\paragraph{Dataset splits.}
The final StreamSum splits contain 2{,}000 training examples, 500 validation examples, and 500 test examples. 
We use the train and validation splits to train \maskdisc{} and one-step repair models, and reserve the test split for final evaluation.

\begin{table*}[t]
\centering
\small
\setlength{\tabcolsep}{4pt}
\begin{tabular}{p{0.18\textwidth}p{0.28\textwidth}p{0.44\textwidth}}
\toprule
\textbf{Stage} & \textbf{Agent / Tool} & \textbf{Purpose} \\
\midrule
Schema generation & Brainstormer agent & Proposes diverse event schemas with one dominant revision point. \\
Seed retrieval & NewsDataHub retriever & Retrieves and deduplicates real-world news articles with high update potential. \\
Timeline construction & Timeline constructor agent & Creates 4--5 update event timelines with early incomplete evidence and later corrective evidence. \\
Timeline hardening & Hardener agent & Improves realism, temporal progression, entity consistency, and revision clarity. \\
Summary construction & Summary constructor agent & Writes a concise full-context reference summary supported by the full context but not fully supported by early context. \\
Support filtering & AlignScore prefix scorer & Measures whether support for the full-context reference summary increases from early prefixes to the full timeline. \\
Validation & Verifier agent & Checks realism, dominant revision quality, support dynamics, and benchmark usefulness. \\
Pruning & Pruner agent & Selects a diverse high-quality subset across event families and revision types. \\
\bottomrule
\end{tabular}
\caption{
StreamSum construction pipeline. 
The benchmark is built with real-world news seeds, agentic timeline synthesis, automatic support-dynamics filtering, and verifier/pruner agents.
}
\label{tab:streamsum_pipeline}
\end{table*}

\subsection{StreamSum Agent Prompts}
\label{app:streamsum_prompts}


\paragraph{Prompts.}
We use separate prompts for schema brainstorming, seed instantiation, timeline construction, timeline hardening, summary construction, verification, and pruning. 
Table~\ref{tab:streamsum_prompts} summarizes the role of each prompt, and full prompt files will be released with the benchmark construction code.

\begin{table*}[t]
\centering
\small
\setlength{\tabcolsep}{4pt}
\begin{tabular}{p{0.18\textwidth}p{0.72\textwidth}}
\toprule
\textbf{Prompt} & \textbf{Instruction Summary} \\
\midrule
Brainstormer & Generate diverse event schemas for revision-aware streaming summarization. Each schema must contain one dominant factual revision point, realistic event grounding, and a clear reason why early updates are misleading. \\
Retriever / Instantiator & Match schemas to retrieved news articles, select temporally ordered updates, and construct candidate early, late, and full contexts. \\
Timeline Constructor & Produce a 4--5 update news-style timeline where early reports are plausible but incomplete and later updates revise the central interpretation. \\
Hardener & Improve realism, temporal progression, entity consistency, information density, and revision clarity while preserving the central event thread. \\
Summary Constructor & Write a concise full-context reference summary that reflects the final event state and requires later updates for support. \\
Verifier & Evaluate candidates using both the timeline text and AlignScore prefix trajectory, scoring coherence, consistency, revision quality, early-context misleadingness, final support, and benchmark usefulness. \\
Pruner & Select a high-quality diverse subset across event families, revision types, and difficulty levels while avoiding near duplicates. \\
\bottomrule
\end{tabular}
\caption{
Summary of StreamSum synthesis prompts. 
Full prompts will be released with the benchmark construction code.
}
\label{tab:streamsum_prompts}
\end{table*}

\paragraph{Full prompts.}
For space, Table~\ref{tab:streamsum_prompts} summarizes the prompt roles. 
The complete prompt files for the brainstormer, retriever/instantiator, timeline constructor, hardener, summary constructor, verifier, and pruner agents will be released with the StreamSum construction code.

\subsection{Example StreamSum Instances}
\label{app:streamsum_example}

Table~\ref{tab:streamsum_examples} shows representative StreamSum examples across domains and revision types. 
Each example contains an early claim that is plausible from the early context and a later correction that changes the support needed for the final summary. 
These examples illustrate the benchmark's focus on evolving evidence rather than static summarization.

\begin{table*}[t]
\centering
\scriptsize
\setlength{\tabcolsep}{3.5pt}
\renewcommand{\arraystretch}{1.08}
\begin{tabular}{p{0.13\textwidth}p{0.14\textwidth}p{0.29\textwidth}p{0.36\textwidth}}
\toprule
\textbf{Domain} & \textbf{Revision} & \textbf{Early claim} & \textbf{Later correction / final state} \\
\midrule

AI / technology & Numeric update &
Only 28\% of Indian respondents say AI ROI has met or exceeded expectations; 38\% of organizations have a formal AI policy. &
Final survey results report 41\% AI ROI satisfaction in India, compared with 25\% globally, and 52\% formal AI policy adoption. \\

Business / finance & Market revision &
CerveCo shares gained approximately 11\% for the week, and Morgan Stanley raised its target to \$4.10. &
CerveCo ultimately rose 16\% for the week, while Morgan Stanley upgraded the stock from Hold to Buy and raised its target to \$4.75. \\

Cybersecurity & Attribution update &
A federal archive breach was attributed to a lone cybercriminal, with no confirmed file downloads. &
Forensics later tied the intrusion to a foreign intelligence-linked group and confirmed that about 2{,}400 documents were exfiltrated. \\

Disaster / safety & Outcome flip &
Two backpackers remained missing and were feared caught in floodwaters; Charleys Creek was forecast to peak at 6.8 meters. &
The backpackers were found alive after sheltering on higher ground, and Charleys Creek peaked lower, at 5.9 meters. \\

Governance & Outcome flip &
The EU--Cameroon Business Forum was scheduled for July 14 in Douala, and the EU accounted for 24\% of Cameroon's trade. &
The forum was postponed indefinitely over logistics and security concerns, and corrected trade figures put the EU share at 31\%. \\

Environment / mining & Timeline change &
TERRA Ambiental was leading the Altamira environmental filing, scheduled for completion by December 2026. &
GeoSur Consultores replaced TERRA after a contract dispute, and the filing deadline moved to June 2027 after regulators requested additional studies. \\

Energy / health & Scientific correction &
A report suggested infant mortality near the Clearwater Nuclear Plant had risen above the state average, while regulators approved a 15-year extension. &
Regulators granted a 20-year extension, but county officials found local infant mortality rates were generally similar to or lower than statewide averages. \\

Healthcare / logistics & Status change &
A cold-chain hub was expected to receive EMA certification within weeks and had processed over 180 shipments. &
After an initial denial, the hub received conditional EMA approval, and auditors confirmed 310 shipments after adding previously unlogged consignments. \\

International / economy & Numeric update &
GRI Analytics projected India's FY27 GDP growth at 6.5\%, with Brent crude around \$92 per barrel. &
GRI revised growth down to 6.1\% after Brent crude surged to \$118 per barrel amid renewed Strait of Hormuz disruptions. \\

Electric vehicles & Numeric update &
The 2028 Aeron eX5 was announced at \$58{,}400 with up to 410 miles of EPA-estimated range. &
The confirmed starting price rose to \$61{,}950, and EPA certification returned a lower top range of 389 miles. \\

Election law & Legal change &
Section 63 allowed ballots lacking official security features, and certificate forgery was no longer a valid petition ground. &
The Supreme Court struck down Section 63, requiring certified security features, while the commission confirmed certificate forgery remains valid grounds for challenge. \\

Entertainment & Outcome flip &
An HBO comedy series was described as a six-part limited series premiering July 10. &
HBO expanded the series to nine episodes and pushed the premiere to August 21 due to post-production delays. \\

\bottomrule
\end{tabular}
\caption{
Representative StreamSum examples. 
Each row shows an early claim and a later correction/final state, illustrating the dataset's controlled revision patterns across domains.
}
\label{tab:streamsum_examples}
\end{table*}

\end{document}